# Detecting AI-Generated Essays in Writing Assessment: Responsible Use and Generalizability Across LLMs

Jiangang Hao

ETS Research Institute, Princeton, NJ 08541

## Abstract

Writing is a foundational literacy skill that underpins effective communication, fosters critical thinking, facilitates learning across disciplines, and enables individuals to organize and articulate complex ideas. Consequently, writing assessment plays a vital role in evaluating language proficiency, communicative effectiveness, and analytical reasoning. The rapid advancement of large language models (LLMs) has made it increasingly easy to generate coherent, high-quality essays, raising significant concerns about the authenticity of student-submitted work. This chapter first provides an overview of the current landscape of detectors for AI-generated and AI-assisted essays, along with guidelines for their responsible use. It then presents empirical analyses to evaluate how well detectors trained on essays from one LLM generalize to identifying essays produced by other LLMs, based on essays generated in response to public GRE writing prompts. These findings provide guidance for developing and retraining detectors for practical applications.

**Keywords: Detection, AI-generated Essay, Writing Assessment, LLMs**

## 1. Introduction

Writing is a foundational literacy skill that underpins human communication and intellectual development. Bazerman (2008) emphasizes its essential role in enabling effective communication, while Powell (2012) calls it "the most important technology in the history of the human species, except for how to make fire." Writing is critical not only for conveying ideas but also for fostering critical thinking and facilitating learning across disciplines (Adler-Kassner & Wardle, 2022; Deane et al., 2008; Weigle, 2002). For this reason, writing assessment plays a central role in educational systems worldwide, providing evidence of language proficiency, communicative effectiveness, and analytical reasoning. Large-scale writing assessments, such as those used for college and graduate admissions and certification, depend heavily on the assumption that essays reflect examinees' original thinking and language abilities.

The rapid advancement of large language models (LLMs) has introduced new challenges to evaluating writing products. Modern LLMs can generate coherent, well-structured, and high-quality essays with minimal user effort. These capabilities have raised significant concerns among assessment programs, educators, and policymakers about the authenticity of student-produced writing. When AI-generated essays are submitted as work intended to reflect human skill, the validity of score interpretations is threatened. These concerns are not merely theoretical, as many surveys and studies now document widespread use of generative AI tools by students to assist their writing assignments, either partially or fully (Cu & Hochman, 2023; Terry, 2023).

In response, there has been rapid growth in methods for identifying AI-generated writing. Some publicly available tools include, but are not limited to, GPTZero (https://gptzero.me), Originality.AI (https://originality.ai), Copyleaks (https://copyleaks.com), Grammarly's AI content detector (https://www.grammarly.com/ai-detector), and Turnitin's AI writing detector

(https://www.turnitin.com/solutions/topics/ai-writing/). Despite the large number of tools, accurate detection of AI-generated text in real-world, open environments remains an unresolved challenge. The absence of controlled conditions makes it difficult to establish trustworthy performance metrics or to ensure consistent behavior across populations, topics, and evolving model versions. A growing body of empirical evidence indicates that many current detection systems frequently misclassify human writing as AI-generated and often fail to reliably differentiate the two across diverse contexts (Harwell, 2023; Sadasivan et al, 2023; Tufts et al., 2025; Weber-Wulff et al., 2023). These limitations raise concerns about fairness, validity, and the responsible use of detection tools in educational or evaluative settings (Hao et al., 2024).

On the other hand, the situation is more tractable when the focus is placed on detecting AI-generated essays within standardized writing assessments. Such assessments are administered under well-defined and proctored conditions. Prompts are fixed, testing time is predetermined, and students must respond within a relatively narrow task structure. These constraints reduce the variety and unpredictability that complicate detection in open environments. Because the prompt set is known, the response space is constrained, and authentic human responses are available at scale, it becomes possible to construct more stable and interpretable performance metrics. In this context, researchers can evaluate detection accuracy, false positive rates, and robustness with greater confidence.

For these reasons, this chapter concentrates on the detection of AI-generated essays in standardized assessment settings. We first review major methodological approaches used for detection, their pros and cons. We then examine the generalizability of detectors across different LLMs to suggest practical detection strategies for the evolving landscape of LLM. Finally, we discuss responsible use of these detectors and outline remaining challenges and identifies promising directions for advancing detection research and practice.

## 2. Detecting LLM-generated Essay

Detecting AI-generated essays is inherently challenging, in part because the very definition of "AI-generated" text is not always clear-cut. Modern LLMs are trained on massive corpora of human-written material, which means they can produce passages that closely resemble authentic human writing, sometimes even reproducing specific patterns or phrases from their training data. The boundary becomes especially ambiguous when the text is short in a constrained context; for instance, determining whether a single word or brief phrase is "AI-generated" is effectively meaningless. A review of detection methods in different scenarios can be found in Grothers et al., (2023). In this chapter, our discussion focuses on a relatively constrained and practically relevant setting: extended essays written in response to prompts from large-scale writing assessments.

### 2.1. Plagiarism Detection Before LLM

Efforts to ensure the authenticity of written work long preceded the emergence of LLM. Historically, concerns focused primarily on traditional forms of plagiarism, e.g., copying or closely paraphrasing text from published sources, online materials, or a student's own prior work. Beginning in the late 1990s and early 2000s, large-scale plagiarism detection systems emerged to address these issues in educational and scholarly contexts. Early systems, such as *Turnitin* (e.g., Culwin & Lancaster, 2001; Maurer, Kappe, & Zaka, 2006), were designed to identify verbatim or near-verbatim overlap between a student's submission and extensive databases of internet content, academic publications, and previously submitted assignments.

Such approaches have also been successfully applied in large-scale writing assessments to detect instances of students copying from prepared or unauthorized source texts (Choi et al., 2024).

Traditional plagiarism detection methods proved effective at detecting traditional forms of plagiarism but were inherently limited: the plagiarized texts must be included in their databases. The rapid rise of transformer-based LLMs represents a fundamental shift. Unlike plagiarism, AI-generated essays are often original compositions with no close textual neighbors in any database. As a result, detection requires a new paradigm, one focused less on identifying copied content but more on recognizing subtle linguistic, stylometric, or behavioral signatures that differentiate human-written from AI-generated texts.

### 2.2 Detectors Used in Writing Assessment

Although a wide range of AI-generated text detectors has been proposed in recent years, the underlying methodological approaches can generally be grouped into a few core paradigms. A seemingly straightforward way to detect AI-generated essays is to ask an LLM (e.g., ChatGPT) whether an essay was produced by AI. However, this approach is highly unreliable and typically yields high false-positive and false-negative rates in practice. In the following sections, we introduce several methodological approaches that have been empirically shown to perform more consistently across a range of contexts and writing conditions.

#### *2.2.1 Supervised Learning Classifier*

The most common approach treats AI-generated essay detection as a supervised learning problem. In this framework, researchers compile corpora of human-written and AI-generated essays and extract features that capture systematic differences between these two categories. Some consider linguistic and stylistic features that, on average, differ between human- and AI-generated text (e.g., Yan et al., 2023; Jiang et al., 2024; Zhong et al., 2024), while others use probabilistic indicators such as perplexity or burstiness to estimate whether a passage is more likely to have been produced by an LLM (e.g., Hao & Fauss, 2024a; Mitchell et al., 2023; Tian et al., 2023). A binary classifier is then trained to distinguish AI-generated from human-produced writing based on these features. Classifiers vary widely and can include logistic regression, random forests, gradient boosting, support vector machines, and more. The performance of the resulting detector depends heavily on the quality of the training data and the representativeness of the features. High detection accuracy has been reported for these approaches across several large-scale writing assessment contexts (Yan et al., 2023; Jiang et al., 2024; Hao & Fauss, 2024a; Zhong et al., 2024).

In addition to explicit feature-based supervised learning approaches, another line of work leverages end-to-end fine-tuning of pretrained LLMs, such as RoBERTa (Liu et al., 2019), using labeled datasets of human-written and AI-generated essays. These models learn discriminative representations directly from raw text during supervised fine-tuning, eliminating the need for manually engineered linguistic or probabilistic features. This approach has shown strong performance, particularly in large-scale, high-stakes writing assessments. For example, Yan et al. (2023) demonstrated that a RoBERTa-based classifier, fine-tuned end-to-end on hundreds of human-written and AI-generated essays, achieved high accuracy for essays from a large-scale writing test.

Both feature-based and end-to-end fine-tuning approaches offer distinct advantages and limitations. Feature-based detectors are generally more interpretable, as the underlying linguistic or probabilistic features can be inspected directly to understand which textual properties differentiate human and AI writing. Because these features often reflect broad stylistic or distributional tendencies, feature-based models also tend to exhibit stronger cross-prompt and cross-model generalizability, making them more robust when encountering new writing prompts and genres. In contrast, end-to-end fine-tuned LLM classifiers typically achieve slightly higher predictive performance, particularly when trained on large, well-matched datasets. However, these gains come at the cost of transparency: the models function largely as black boxes, making it challenging to explain their decisions or diagnose failure modes. Moreover, because end-to-end models learn highly specific representations tied to the training distribution, their generalizability may degrade sharply when applied to essays from new prompts, genres, or unseen LLMs.

*2.2.2 Watermarking*

Instead of inferring whether an essay has been generated by an AI system, another approach is to add a watermark, e.g., a statistical signature, to AI-generated texts, which can later be identified using specialized algorithms (Kirchenbauer et al., 2023). A common strategy biases the model's sampling process toward a designated subset of tokens, producing subtle distributional signals that can later be detected. Because the watermark is introduced during generation rather than inferred afterward, this approach offers clear provenance and avoids reliance on handcrafted features or large supervised datasets.

However, watermarking has several major limitations that restrict its real-world utility. The watermark signal is highly fragile: even modest paraphrasing, editing, or re-generation through another model can weaken or erase it. Watermarking also requires cooperation from LLM developers and cannot be retrofitted to existing or open-source systems, making broad adoption unlikely. In adversarial or open-world environments, an attacker can use non-watermarked models, remove the signal, or apply transformations that invalidate detection.

Overall, watermarking can provide positive evidence that a text was generated by a compliant LLM, but it cannot reliably verify that a text is human-written when no watermark is found.

*2.2.3 Writing Process*

Beyond linguistic and probabilistic features, writing process–based features offer another promising direction for detecting AI-generated or inauthentic text. In controlled digital writing environments, keystroke dynamics, revision histories, and timing information can be captured unobtrusively. Human writers typically exhibit natural, irregular behavioral patterns, including pauses, bursts of typing, backtracking, and revisions. These traces reflect genuine cognitive and compositional processes.

When an AI-generated essay is copied and pasted into a writing interface, or when a writer manually transcribes an externally generated text, these natural behavioral signatures are largely absent or appear in highly altered form. As a result, incorporating process-based indicators can substantially enhance detection performance, especially in proctored or computer-based assessment settings where detailed interaction logs are available.

A growing body of evidence supports the utility of such process data. Keystroke dynamics has been shown to function as a reliable biometric measure for identifying repeated writers (Choi et al., 2021), and recent studies demonstrate that process-based features effectively distinguish authentic human writing from copywriting or AI-generated responses (Deane et al., 2025; Jiang et al., 2025). Together, these findings highlight the potential of writing process data as a robust complement to text-based approaches in high-stakes assessment contexts.

*2.2.4 Similarity Matching*

The methods described in the preceding subsections perform well when an essay is produced entirely by an AI system with little to no human editing layered on top of an AI-generated draft. In real-world settings, however, this assumption rarely holds. Human writers often modify, extend, reorganize, or lightly edit AI-generated content, producing hybrid responses that blend human and machine contributions. In these situations, detecting AI involvement becomes extremely challenging, and the very definition of what counts as "AI-generated" becomes blurred. At present, no generally applicable method can reliably identify such mixed-origin texts.

A partial exception arises in standardized writing assessment settings where prompts are fixed and tightly constrained. In these contexts, a similarity-based approach can be used to detect potential AI influence. Because the set of prompts is known in advance, one can generate a large pool of AI-generated essays for each prompt (e.g., 200 per prompt). Human-submitted essays can then be compared against this pool to identify cases where substantial overlap suggests reliance on AI-generated material. Hao and Fauss (2024b) developed such a detector, GPTCollider, which has shown promising results in practical applications. Figure 1 illustrates an example of the overlap between a human-submitted essay and an AI-generated essay identified through this method.

**Figure 1**. Overlapping of human-submitted and AI-generated essay segments.

Human-submitted

According to the given statement,as it profess that the beginner is more likely than the expert to make important contributions in any field of inquiry is a controversial one. While it is true that beginners may bring fresh perspectives and ideas to a field, it is also true that experts have a wealth of knowledge and experience that can lead to significant breakthroughs. In this essay, I will argue that while beginners can make important contributions, it is ultimately the experts who are more likely

AI-generated

The statement that beginners are more likely than experts to make important contributions in any field of inquiry is a complex one that requires careful consideration. While it is true that beginners may bring fresh perspectives and ideas to a field, experts also have a wealth of knowledge and experience that can lead to significant contributions. In this essay, I will argue that while beginners can make important contributions, experts are more likely to do so due to their extensive

It is important to emphasize that this approach is feasible only when the number of prompts is small and predetermined, which is typical in standardized writing assessments. In open-ended writing environments or assessments with a large or dynamic set of prompts, it is impractical to generate sufficiently comprehensive AI-produced essay pools in advance, and the method becomes ineffective.

## 3 Generalizability of Detectors Across LLMs

As increasingly capable LLMs continue to emerge, the landscape of AI-generated writing is evolving at a rapid pace. Detectors developed for earlier generations of LLMs often struggle when faced with the stylistic diversity, improved fluency, and reduced errors of newer systems. This constant stream of more powerful LLMs means that detection is not a one-time problem but an ongoing challenge: methods that perform well today may degrade substantially as models become more sophisticated or adopt different training strategies. Consequently, it is important to evaluate how well detectors trained on one LLM generalize to others and to understand the extent to which detection techniques remain effective as the underlying generative models evolve. By studying cross-model performance, we can assess the robustness of detection approaches, identify vulnerabilities introduced by newer models, and develop guidance for designing detectors that are adaptable, forward-compatible, and capable of keeping pace with the rapid development of generative AI technologies.

Based on two publicly available writing prompts from the Graduate Record Exam (GRE), Zhong et al. (2024) investigated the extent to which detectors trained on essays generated by one LLM generalize to essays produced by other models, focusing on LLMs available through early 2024. Extending this line of work, we evaluate cross-model performance for a broader set of more powerful GPT-family models released in 2024 and 2025, including GPT-4, GPT-4o, GPT-o1, GPT-o3-mini, GPT-o4-mini, and GPT-5. Using the same two writing prompts (see Appendix for details), we generated 200 essays per prompt for each LLM (400 per model). We also sampled 100 human-written essays per prompt, yielding 200 total human responses. From these, 100 human essays were selected as a universal test set for evaluating every detector. For each LLM, we then selected 100 of its generated essays to pair with 100 human essays, creating a balanced testing set specific to that model. The remaining data (300 LLM-generated essays and 100 human essays) served as the training set for that model's detector.

Zhong et al. (2024) demonstrated that perplexity-based features are highly effective for detecting AI-generated essays. Perplexity measures how likely a sequence of words is under a given language model and thus provides a useful signal for distinguishing human- and machine-authored text. Because most LLMs examined in this study do not support direct perplexity computation, we used GPT-2 as a common reference model to compute perplexity for all essays. In this context, perplexity represents how well each essay aligns with GPT-2's language distribution. For each essay, we extracted a set of perplexity-derived features, including the overall essay perplexity and the mean, median, minimum, maximum, and the 10th through 90th percentiles of the sentence-level perplexity distribution.

For each LLM, we fine-tuned a Gradient Boosting Machine using four-fold cross-validation on the model-specific human–AI training data, optimizing the detector to distinguish essays generated by LLMs from those written by humans. The best classifier for each model was then applied to the balanced testing sets of all LLMs (including its own), and the Area Under the ROC Curve (AUC) was computed as the primary evaluation metric. AUC provides a threshold-free measure of discriminative performance by summarizing the ROC curve, which plots true positive rate against false positive rate into a single value between 0 and 1. A value of 1 indicates perfect separation, 0.5 reflects chance performance, and values below 0.5 indicate systematic misclassification. Because AUC is insensitive to class imbalance and does not depend on a single decision threshold, it offers a robust basis for comparing cross-model generalizability. In addition to the model-specific detectors, we also trained a unified detector using the combined training data from all LLMs; we refer to this model as GPT-all. The resulting cross-LLM detection AUC matrix is shown in Figure 2.

**Figure 2.** AUC-based detection performance for classifiers trained and tested on essays from different LLMs.

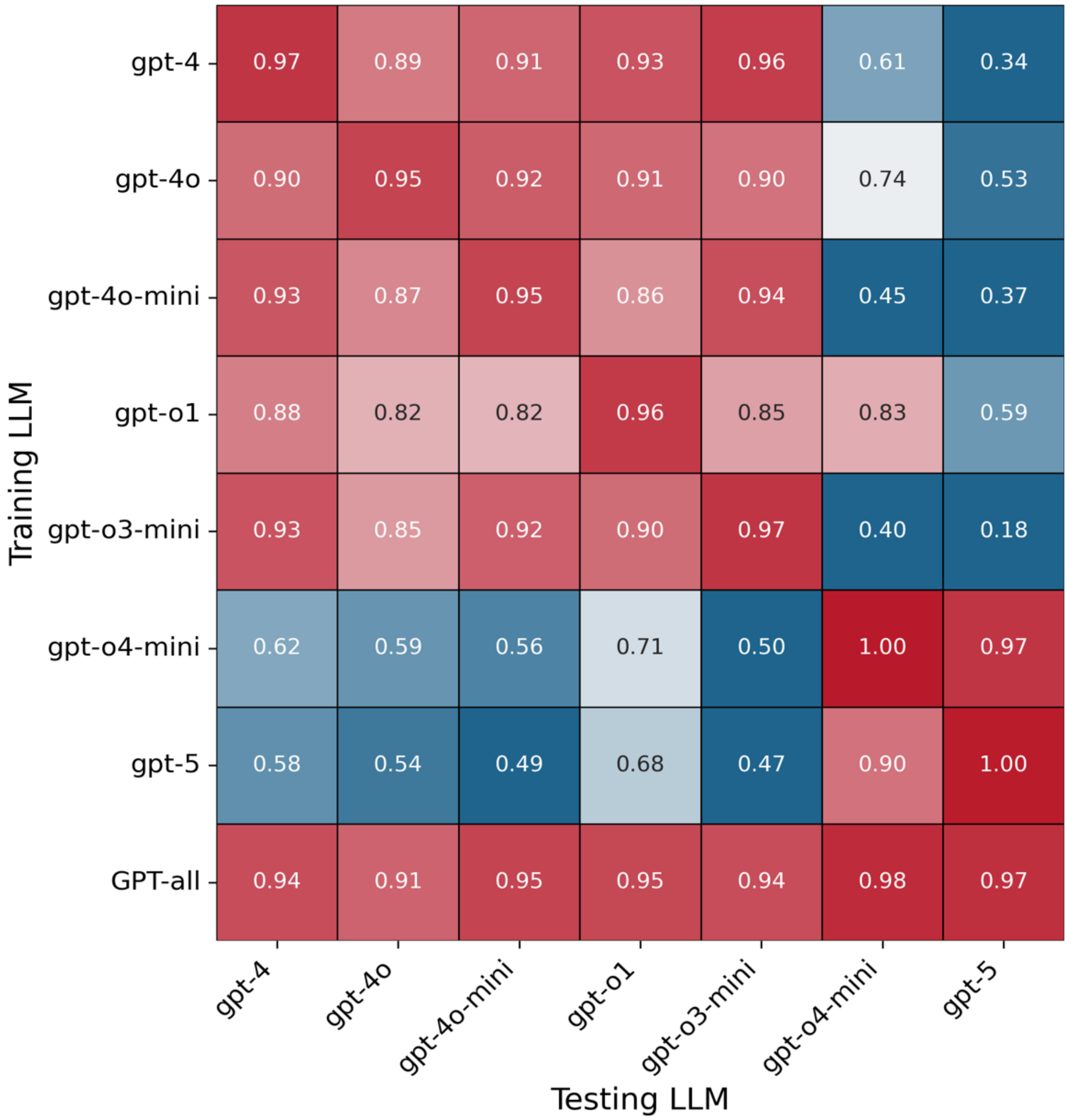

The cross-LLM AUC analysis highlights several important patterns in how detectors trained on different GPT-family models generalize across a range of related LLMs. A consistent observation is that detectors perform very well when evaluated on the same model used during training. Across all seven LLMs, the diagonal values in the heatmap remain high, indicating that each detector successfully captures the unique generative characteristics of its corresponding model. This suggests that when the training and testing distributions match, distinguishing human-written text from LLM-generated text remains relatively easy for the feature-based GBM classifier.

Beyond within-model detection, the results show substantial cross-LLM generalizability among several GPT models. In particular, GPT-4, GPT-4o, GPT-4o-mini, GPT-o1, and GPT-o3-mini consistently achieve AUC values above 0.8 when detecting each other's outputs, with many values exceeding 0.9. This cluster exhibits similar stylistic patterns or shared statistical regularities, enabling detectors trained on one model to transfer effectively to the others. The presence of strong bidirectional generalization suggests that these models, despite differences in size or optimization, occupy a relatively coherent region of the GPT generative space. Detectors trained on any one of them acquire information that transfers well to the others.

In contrast, the models GPT-o4-mini and GPT-5 show a divergent pattern. These two LLMs display relatively strong mutual generalizability, with detectors performing well when trained on one and tested on the other. However, both models exhibit poor generalization across the rest of the GPT family, and detectors trained on other models generally struggle to detect text generated by GPT-o4-mini and GPT-5. This pairing suggests that these two LLMs share certain stylistic or structural features that set them apart from the earlier GPT-4 and GPT-o series. However, when we train a detector by combining training data from all LLMs (GPT-all), the resulting detector can detect essays generated by each LLM very well.

The fact that such variation exists within the GPT family of models underscores a broader challenge for AI-generated text detection. If generalizability varies this substantially within a single corporate lineage of LLMs, it is reasonable to expect even larger gaps when detectors are applied to outputs from LLMs developed by other companies, such as Google's Gemini or Anthropic's Claude (Zhong et al., 2024). As such, a working strategy should be to include training data generated from all LLMs (e.g., the GPT-all row in the Figure 2).

Overall, the results highlight both strengths and vulnerabilities of current detection approaches. Within-model detection remains strong, and several GPT models form a well-behaved cluster with high mutual generalizability. At the same time, the emergence of isolated clusters like gpt-o4-mini and gpt-5 suggests that robust AI text detection will increasingly require training across multiple LLM families to capture the growing diversity of generative behaviors.

**4. Responsible Use**

In the preceding sections, we described how different types of detectors for AI-generated essays function and reviewed their respective strengths and limitations. A key message is that no detector can be expected to perform consistently well across all scenarios. Detection models may exhibit unequal error rates across demographic groups, second-language writers, or individuals with atypical writing styles. Therefore, the use of these detectors, especially in high-stakes educational or evaluative settings, must be approached with caution. In this section, we summarize several critical issues that users should be aware of to ensure responsible use of these detectors (Hao et al., 2024):

First, no AI-generated text detector is perfect. All detectors produce false positives and false negatives, so users should interpret their output with caution and with an understanding of the detector's performance metrics. Because every detector is trained on a limited dataset, its reliability decreases when applied to scenarios that differ from the training conditions. In practice, no existing detector performs consistently well across all types of writing tasks or contexts.

Second, text length also plays a critical role in detector accuracy. Performance drops sharply for short responses because there is insufficient linguistic or stylistic information to differentiate human- from AI-written text. In the extreme case of a single word, the distinction becomes meaningless, since there is no principled way to determine whether it originated from a human or an AI system.

Third, most detectors struggle when an essay is jointly produced by humans and AI. Once human edits are layered onto AI-generated content, the boundary between the two becomes blurred, making reliable identification extremely difficult. Users must also be aware that some detectors may exhibit bias against certain demographic groups, underscoring the need for careful evaluation before using detector results in any consequential decision-making.

Finally, different detectors may produce conflicting judgments for the same essay, a point that students or writers may raise when disputing accusations of AI use. For this reason, whenever possible, detector outputs should be supplemented with additional sources of information. Writing process data, such as keystroke logs or video recordings of the writing session, can provide valuable contextual evidence that strengthens or clarifies conclusions drawn from text-based detection alone.

On the other hand, despite the many issues, we should also avoid overreacting. For example, Liang et al., (2023) reported, based on a small dataset, that many publicly available detectors exhibit a higher false-positive rate when flagging essays written by non-native English speakers. This finding, together with some less systematic observations in practice, led to voices for banning the detectors. However, in later studies, Jiang et al. (2024a; 2024b), based on a large dataset, showed that carefully developed detectors do not necessarily bias against non-native English speakers. As such, general policy discussion should be based on balanced evidence and sufficiently consider both the pros and cons of using AI-generated text detectors in different contexts. We do not ban cars because of car accidents; instead, we implement safety measures and regulations to minimize risks. Similarly, with AI text detectors, we should focus on responsible use, training, and awareness rather than outright bans.

More broadly, responsible use of AI-generated text detectors requires institutional consensus rather than ad hoc decisions by individual instructors. Institutions should develop shared guidelines that clearly define appropriate use cases, limitations, and safeguards, and should communicate these expectations transparently to students. In instructional settings, detector use should be complemented by evaluation designs that combine take-home assignments with in-class or proctored writing tasks, allowing instructors to examine consistency across contexts. Such design-based approaches reduce reliance on detector outputs alone and provide a more robust basis for interpreting student work.

## 5 Future

In this chapter, we summarized common methods for detecting AI-generated essays in writing assessments. Our discussion focused primarily on fully AI-produced texts, though we also noted that certain text-matching approaches may extend to identifying hybrid human–AI writing in constrained situations. As AI-assisted writing becomes commonplace in everyday communication, the nature of writing itself is undergoing a significant transformation. Increasingly, people rely on AI tools to generate ideas, refine arguments, and correct grammar, usage, and mechanics. This evolution raises important questions about how AI will reshape human writing and, in turn, which aspects of writing should remain central in evaluation. If AI tools can reliably handle surface-level features such as grammar and mechanics, it may be necessary to reconsider the weight these dimensions should carry in scoring. At the same time, the increasing integration of AI into the writing process will make it even more challenging to distinguish between human-written and AI-generated text. For this reason, detection models must be routinely re-evaluated to monitor performance drift and updated as new LLMs emerge or as patterns in human writing evolve.

This chapter has focused on the detection of AI-generated text in writing assessment. Many of the considerations and guidelines for the responsible use discussed here, however, extend beyond text and are equally relevant to the detection of AI-generated content in other modalities, such as audio, video, and images. For example, deepfake detection raises parallel challenges related to imperfect accuracy, context dependence, bias, and the risk of misuse in high-stakes decisions. Across modalities, no detector can provide definitive proof of AI involvement, and responsible use requires careful interpretation, transparency, and the integration of multiple sources of evidence. As generative AI continues to advance across media, the principles of cautious deployment, institutional consensus, and informed human judgment will remain central to the responsible use of detection technologies.

## Appendix

The two prompts used to instruct the LLMs to generate essays are listed below. They are drawn directly from publicly available GRE writing prompts, with an additional one-line requirement on the length, "Please keep the response to about 500 words."

*Prompt 1*: Governments should not fund any scientific research whose consequences are unclear. Write a response in which you discuss the extent to which you agree or disagree with the recommendation and explain your reasoning for the position you take. In developing and supporting your position, describe specific circumstances in which adopting the recommendation would or would not be advantageous and explain how these examples shape your position. Please keep the response to about 500 words.

*Prompt 2*: To understand the most important characteristics of a society, one must study its major cities. Write a response in which you discuss the extent to which you agree or disagree with the statement and explain your reasoning for the position you take. In developing and supporting your position, you should consider ways in which the statement might or might not hold true and explain how these considerations shape your position. Please keep the response to about 500 words